\documentclass[conference]{IEEEtran}
\IEEEoverridecommandlockouts
\usepackage{float}
\usepackage{cite}
\usepackage{amsmath,amssymb,amsfonts}
\usepackage{algorithmic}
\usepackage{graphicx}
\usepackage{textcomp}
\usepackage{xcolor}
\usepackage{hyperref}
\usepackage{comment}

\hypersetup{colorlinks=true,urlcolor=blue, citecolor=black}

\def\BibTeX{{\rm B\kern-.05em{\sc i\kern-.025em b}\kern-.08em
    T\kern-.1667em\lower.7ex\hbox{E}\kern-.125emX}}

\makeatletter
\newcommand{\linebreakand}{%
  \end{@IEEEauthorhalign}
  \hfill\mbox{}\par
  \mbox{}\hfill\begin{@IEEEauthorhalign}
}
\makeatother

\begin{document}

\title{Forecasting Pressure Of Ventilator Using A Hybrid Deep Learning Model Built With Bi-LSTM and Bi-GRU To Simulate Ventilation\\
}

\author{\IEEEauthorblockN{\textsuperscript{} Md. Jafril Alam}
\IEEEauthorblockA{\textit{Department of Computer Science and Engineering} \\
\textit{Khulna University of Engineering \& Technology}\\
Khulna, Bangladesh \\
jafrilalamshihab.kuetcse@gmail.com}
\and
\IEEEauthorblockN{\textsuperscript{} Jakaria Rabbi}
\IEEEauthorblockA{\textit{Department of Computer Science and Engineering} \\
\textit{Khulna University of Engineering \& Technology}\\
Khulna, Bangladesh \\
jakaria\_rabbi@cse.kuet.ac.bd}
\linebreakand 
\IEEEauthorblockN{\textsuperscript{} Shamim Ahamed}
\IEEEauthorblockA{\textit{Department of Computer Science and Engineering} \\
\textit{Khulna University of Engineering \& Technology}\\
Khulna, Bangladesh \\
shamim.pavel21@gmail.com}
}

\maketitle

\begin{abstract}
A ventilator simulation system can make mechanical ventilation easier and more effective. As a result, predicting a patient's ventilator pressure is essential when designing a simulation ventilator. We suggested a hybrid deep learning-based approach to forecast required ventilator pressure for patients. This system is made up of Bi-LSTM and Bi-GRU networks. The SELU activation function was used in our proposed model. MAE and MSE were used to examine the accuracy of the proposed model so that our proposed methodology can be applied to real-world problems. The model performed well against test data and created far too few losses. Major parts of our research were data collection, data analysis, data cleaning, building hybrid Bi-LSTM and Bi-GRU model, training the model, model evaluation, and result analysis. We compared the results of our research with some contemporary works, and our proposed model performed better than those models.
\end{abstract}

\begin{IEEEkeywords}
Ventilator pressure , Bi-LSTM , Bi-GRU , Deep Learning , Time Series Data
\end{IEEEkeywords}

\section{Introduction}
In medicine and medical transportation systems, many diseases necessitate a ventilator. The efficiency of a mechanical ventilator was proved in the 1950s during the poliomyelitis pandemic\cite{me-vent}. If a person cannot breathe adequately due to hypoxemia, hypercapnia, or respiratory failure, the ventilator can help flow air in and out of the lungs.

Another application of a ventilator is if anyone is going to have surgery with general anesthesia, they need it for proper breathing. However, it is very hard to maintain mechanical ventilator services for all patients during a pandemic because it is time-consuming, costly, and less effective. However, machine learning can help to predict and select appropriate pressure automatically.

Machine learning-based simulation ventilators can anticipate ventilator pressure for any patient, and mechanical ventilators can use that information to offer services to patients. In addition, deep learning models like LSTM, GRU, Bi-LSTM, and Bi-GRU excel with time series prediction and sequence modeling.
\newline
In this research paper, the next sections contain the following parts: related works, motivation, goals of this research, theory and proposed methodology, experiment, experimental result, conclusion, and limitations. Theory and methodology contain two parts: theory of different neural network models for time series analysis and a brief discussion of our proposed methodology.

\section{Related Works}
 Mong Yang et al.\cite{mo} suggested a Bi-LSTM-based model for financial time series data prediction.Rong Liang et al.\cite{rong} suggested a model for estimating mine gas concentration based on Bi-GRU. A hybrid LSTM-GRU model was developed by Noureen Zafar et al.\cite{zafar} to predict city traffic speed. A deep learning based model for  pressure prediction was proposed by NP Sable et al.\cite{NP} where LSTM was used.

\section{Motivation and Goal}
Covid-19, Pneumonia, and many other diseases make breathing difficult and result in a lack of oxygen in the blood. Therefore, a ventilator is required to maintain a normal oxygen rate. However, manually handling mechanical ventilators is expensive and inefficient; thus, using the manual ventilation process is critical. Ventilation simulation has the potential to reduce both costs and difficulties. As a result, Google and Princeton University organized a Kaggle research challenge to find effective machine learning-based models to simulate breathing and control mechanical ventilators. Our goal was to create a machine learning-based model to assist in the control of mechanical ventilators. Our system will predict the required ventilator pressure for a certain patient over a time period.

\section{Theory and Methodology}

\subsection{Different Deep Learning Algorithm for time series data }

\subsubsection{LSTM}
Long short-term memory(LSTM)\cite{lstm} is a recurrent neural network type extensively employed in time series prediction and natural language processing. This neural network can learn the sequencing process, whereas basic artificial neural networks cannot. The LSTM is made up of three main gates: input, output, and forget gate. Input gate, output gate, and forget gate are used to accept input, generate output, and forget the memory. Long-term memory and short-term memory are used as memory cells in LSTM.
Current data, previous hidden state, and internal state are used as input in an LSTM. The values of gates, current hidden state, and current state are computed following corresponding equations\cite{l-eqn}.

\begin{equation} \label{eqn}
f_t = sigmoid(W_{fh}[h_{t-1}] , W_{fx}[x_t] , b_f)
\end{equation}
\begin{equation} \label{eqn}
i_t = sigmoid(W_{ih}[h_{t-1}] , W_{ix}[x_t] , b_i)
\end{equation}
\begin{equation} \label{eqn}
c_t = tanh(W_{ch}[h_{t-1}] , W_{cx}[x_t] , b_c)
\end{equation}
\begin{equation} \label{eqn}
o_t = sigmoid(W_{oh}[h_{t-1}] , W_{ox}[x_t] , b_o)
\end{equation}
\begin{equation} \label{eqn}
h_t = o_t * tanh(c_t)
\end{equation}

Here, $f_t$ , $i_t$ , $c_t$ , $o_t$ , and $h_t$ represent the result of the forget gate,the input gate, candidate vector, the result of output gate, and the memory of LSTM, respectively. $W_{fh}$ , $W_{fx}$ , $W_{ih}$ ,$W_{ix}$, $W_{oh}$ , $W_{ox}$ represent weight metrics . $b_f$ , $b_i$ ,$b_c$ ,$b_o$ indicate bias corresponding to different gates . tanh and sigmoid are two nonlinear activation functions.

\subsubsection{Bi-LSTM}
Bi-LSTM is a type of RNN, which is the expended architecture of LSTM. It overcame the problem of the one-directional information capture flow of LSTM. Bi-LSTM can capture information from both past and present flow. It contains a backward hidden layer and a forward hidden layer. Fig-1 shows the basic structure of bi-directional long short-term memory\cite{b_pic}. "In" and "out" denotes input and output respectively.

\begin{figure}
    \centering
    \includegraphics[height=6cm ]{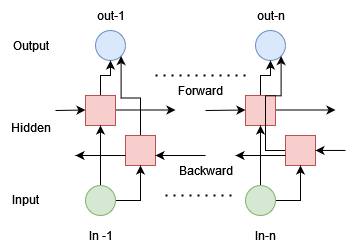}
    \caption{Basic structure of layers of Bi-LSTM}
    \label{fig:my_label}
\end{figure}

\subsubsection{GRU}
GRU, or gated recurrent network, is an RNN proposed by Cho et al.\cite{cho} .It does not have extra memory cells but contains a gating unit capable of modulating information flow inside the unit\cite{gru2}. GRU is implemented using the following equations\cite{bgr1}\cite{bgr2}:
\begin{equation} \label{eqn}
r_t = sigmoid(W_r . [h_{t-1} , x_t])
\end{equation}
\begin{equation} \label{eqn}
z_t = sigmoid(W_z . [h_{t-1} , x_t])
\end{equation}
\begin{equation} \label{eqn}
\widetilde{h_t} = tanh(W_{\widetilde{h_t}} . [r_t * h_{t-1},x_t])
\end{equation}
\begin{equation} \label{eqn}
h_t = (1-z_t) * h_{t-1} + z_t * \widetilde{h_t}
\end{equation}
\begin{equation} \label{eqn}
o_t = sigmoid(W_o . h_t)
\end{equation}

Here, $r_t$ , $z_t$ , $h_t$ , $\widetilde{h_t}$ denotes the reset gate, the update gate, the hidden state, and the candidate vector, respectively, in t timestamp. $W_r$ , $W_z$ , $W_o$ are the weight metrics .

\subsubsection{Bi-GRU}
A bi-directional gated recurrent unit(Bi-GRU) is composed of backward and forward GRU. Both backward and forward GRU is used for obtaining future information and memorizing past information, respectively. Fig -2 shows the basic structure of Bi-GRU\cite{bgr1}.

\begin{figure}[H]
    \centering
    \includegraphics[height=6cm]{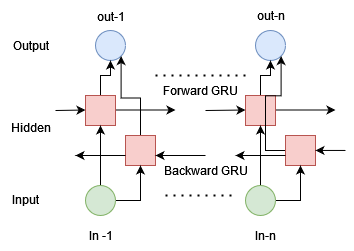}
    \caption{Basic structure of layers of Bi-GRU}
    \label{fig:my_label}
\end{figure}

\subsection{Proposed Methodology}
Fig-3 depicts our proposed methodology for predicting ventilator pressure. Working with data, modeling, and model evaluation are two essential aspects of the methodology. The methodology's first two steps are data collection and preprocessing. The next step is to construct the proposed DNN model. The following steps are model training, testing, and evaluation.

\begin{figure}[!h]
\centering
\includegraphics[width=8cm]{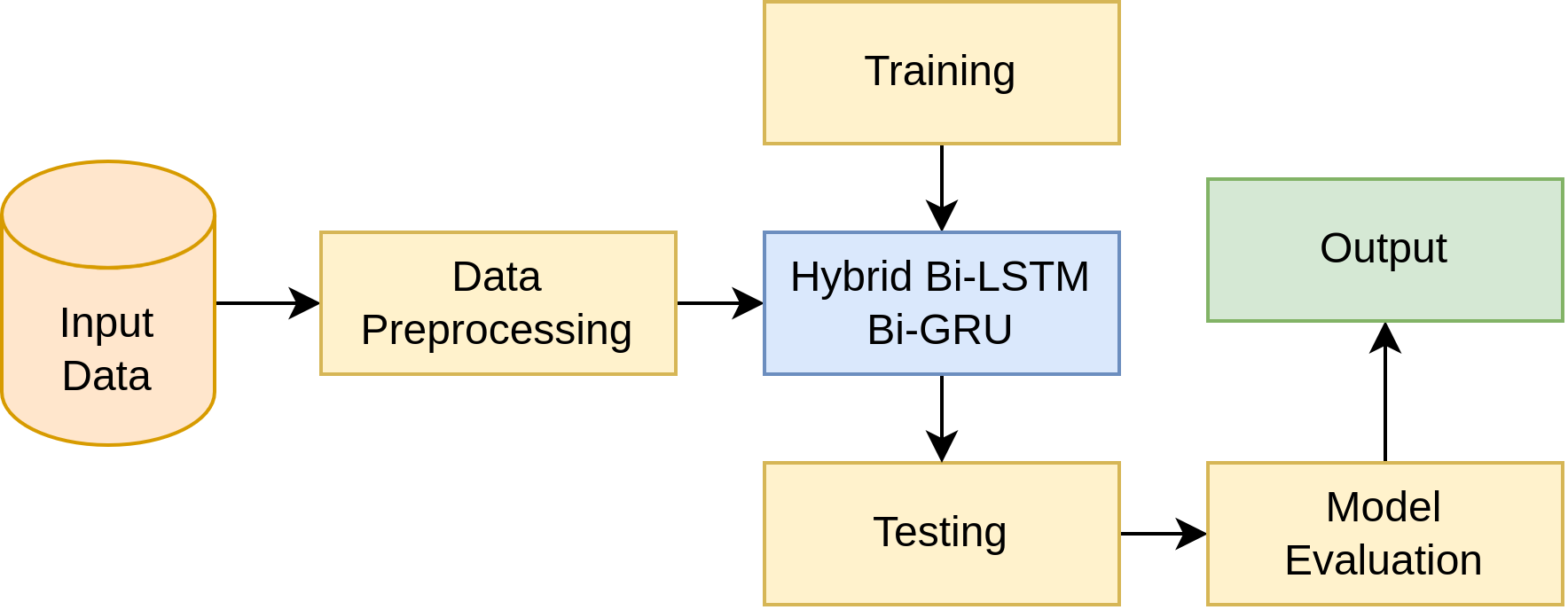}
\caption{Proposed methodology to forecast pressure of a ventilator}
  \label{fig1}
\end{figure}

\subsubsection{Proposed model}
Fig-4 depicts our proposed model for predicting ventilator pressure. This model contains seven Bi-LSTM layers, five Bi-GRU levels, four multiply layers, and five batch normalization layers. The input layer is connected with a Bi-LSTM layer, which is also connected to another LSTM layer. A Bi-LSTM and a Bi-GRU layer receive the previous layer's output. The multiply layer receives one Bi-LSTM and one Bi-GRU of the same number of units as input and outputs a single tensor. In our model, the multiply layer exists four times. Batch normalizing employs the outputs of multiply layers. A dense layer is used as the output layer. The model uses the SELU activation function, which produces the model's output using results generated from Bi-LSTM and Bi-GRU. SELU activation function was proposed by G. Klambauer et al.\cite{selu1} and formulated as follows:
\begin{equation} \label{eqn}
SELU(x) = \lambda * x  \  if  \ x > 0
\end{equation}
\begin{equation} \label{eqn}
SELU(x) = \lambda * (\alpha * e^x - \alpha)  \ if \ x <= 0
\end{equation}

\begin{figure*} [!h]
\centering
\includegraphics[width=16cm, height = 7cm]{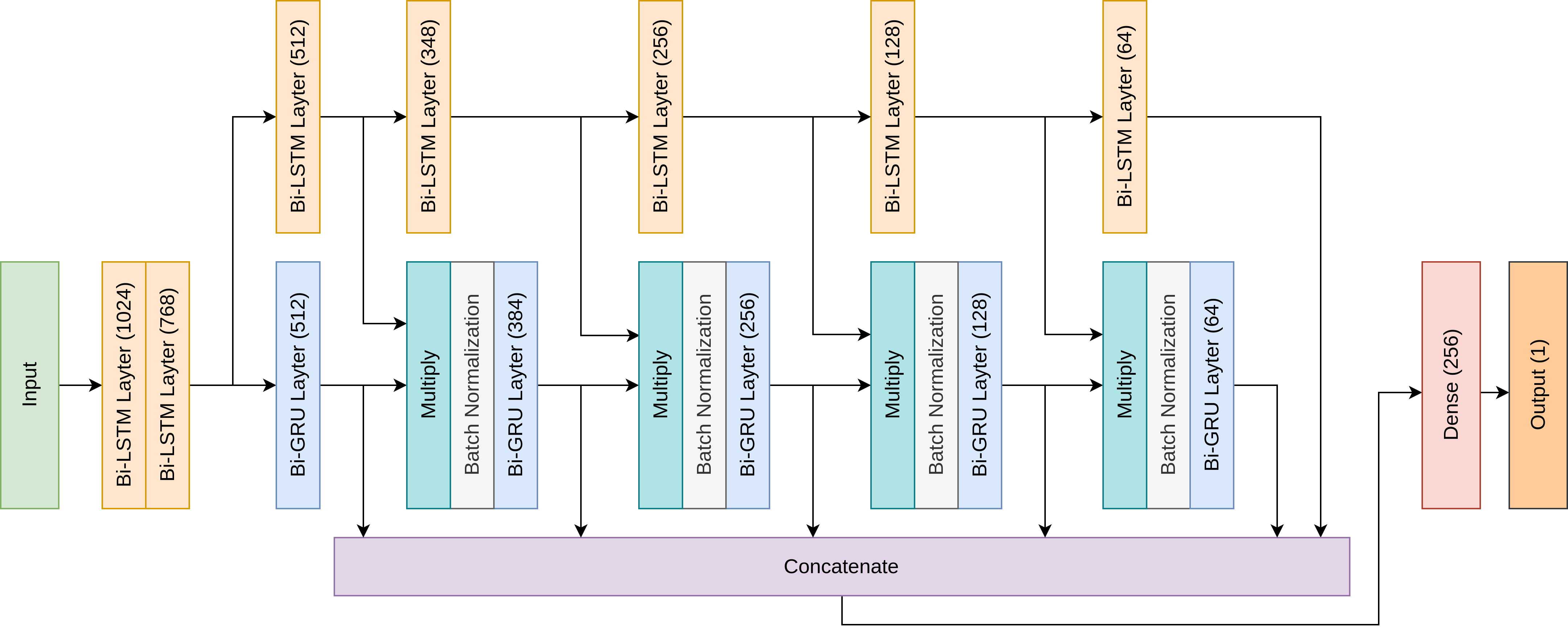}
\caption{Proposed hybrid Bi-LSTM and Bi-GRU based model to forecast pressure of a ventilator}
  \label{fig1}
\end{figure*}

where:
\begin{itemize}
  \item $x$ is the input
  \item $ \alpha $ = 1.6732632423543772848170429916717
  \item $ \lambda $ = 1.0507009873554804934193349852946
\end{itemize}

$ \alpha $ and $ \lambda $ are two constants.

\section{Experiment}

\subsection{Data}
The datasets used in our study were obtained from Kaggle, which is located in \href{https://www.kaggle.com/competitions/ventilator-pressure-prediction/data}{https://www.kaggle.com/competitions/ventilator-pressure-prediction/data}. 
Google Brain published this dataset, and it contains time series data. The numbers of data in train and test sets data are 6036000 and 4024000, respectively. The data was generated by connecting a fully-open supply-chain resilient pressure control ventilator to an artificial lung. This type of ventilator includes modules for the GUI, controller, alarm, common, and IO. The artificial lung is small and can stimulate a patient's condition. The table below summarizes information regarding important features and target columns.

\begin{table}[H]
\centering
\caption{Description of features of the dataset}\label{tab1}
\begin{tabular}{|l|l|}
\hline
Feature & Description\\
\hline
R & It indicates restriction of the airway(in cmH2O/L/S)\\
\hline
C  & It indicates tractablity of the lung (in mL/cmH2O) \\
\hline
time-step & It indicates time step\\
\hline
$u_{in}$ & Control input for the inspiratory solenoid valve \\
\hline
$u_{out}$ &  Control input for the exploratory solenoid valve\\
\hline
pressure & Pressure of air(in cmH2O) , Target column\\
\hline
\end{tabular}
\end{table}

\subsection{Data Analysis and Preprocessing}
Data processing and analysis are critical in machine learning tasks to achieve more accurate and efficient results. As a result, we investigated the pressure in the training dataset, where the minimum and highest pressures are -1.8957 and 64.8209 cmH2O, respectively. Missing values are imputed as part of data processing to produce a more accurate result.
The data were scaled using robust scaling, a well-known technique. It is written as follows\cite{robust}:

\begin{equation} \label{eqn}
Robust Scaling(X[i]) = \frac{X[i] - median(X)}{IQR(X)}
\end{equation}
where:
\begin{itemize}
  \item $X$ is the data
  \item $IQR$ is the inter-quartile range
\end{itemize}

\subsection{Train the model}
The model was trained with the training dataset after data preprocessing and model selection. To implement the program for our proposed system, we used Keras and Python. During training, some of the parameters were tuned, and the critical parameters were called. The table below summarizes key parameters and gives an overview of the model.

\begin{table}[H]
\centering
\caption{Important parameters and model summury during training of the model}\label{tab1}
\begin{tabular}{|l|l|}
\hline
Total params & 54,733,569\\
\hline
Activation function & SELU\\
\hline
Batch size & 512 \\
\hline
Main layers & Bi-GRU , Bi-LSTM \\
\hline
Optimizer &  Adam \\
\hline

\end{tabular}
\end{table}

\subsection{Evaluation metrics}
Performance metrics are critical for justifying and examining any machine learning model since the accuracy of a machine learning model is critical for applying that model practically. As a result, mean absolute error (MAE) and mean squared error (MSE) was employed to assess the proposed model's error. MAE can be expressed mathematically as:

\begin{equation} \label{eqn}
MAE = {\frac{1}{N}\sum_{k=1}^{N}|y_{pr[k]} - y_{ac[k]}|}
\end{equation}
MSE is formulated as :
\begin{equation} \label{eqn}
MSE = {\frac{1}{N}\sum_{k=1}^{N}(y_{pr[k]} - y_{ac[k]})^{2}}
\end{equation}
where:
\begin{itemize}
  \item $y_{pr}$ is the predicted output
  \item $y_{ac}$ is the actual output
  \item $N$ is the total number of samples
\end{itemize}

\section{Experimental Result}
The results and related graphs were saved after the final execution and training of models for ventilator pressure prediction. The MAE and MSE of our proposed model were 0.145 and 0.094 respectively. 

The graph of pressure(in cmH2O) vs. time is shown in Fig-5. This graph depicts the projected pressure at a specific time-step, with the dotted red line representing predicted pressure and the rest line representing the ventilator's actual pressure.
\begin{figure} [H]
\centering
\includegraphics[width=9cm , height = 5cm ]{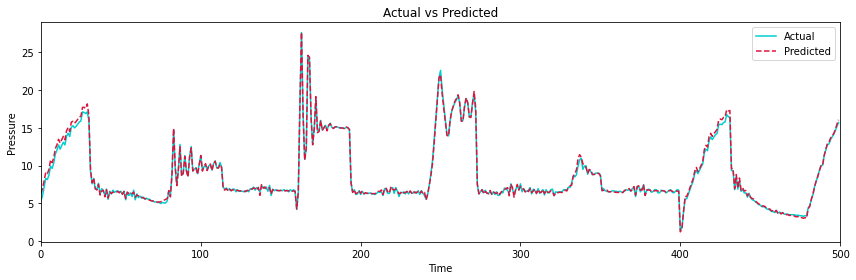}
\caption{Graph of predicted pressure vs time-step}
  \label{fig1:my_label}
\end{figure}
From the fig-5 it is clear that , difference between the actual result and predicted result is too little which indicates good performance of the proposed model. Fig-6 depicts the MSE vs. Epochs graph, whereas Fig-7 depicts the MAE vs. Epochs graph. The graph demonstrates that MAE and MSE decreased significantly after a few epochs. From Fig-6 and ,7 it is also clear that , after few epochs, error of the proposed model decreased significantly which denotes the good accuracy and higher acceptability of the model.

\begin{figure}[H]
    \centering
    \includegraphics[height=6cm]{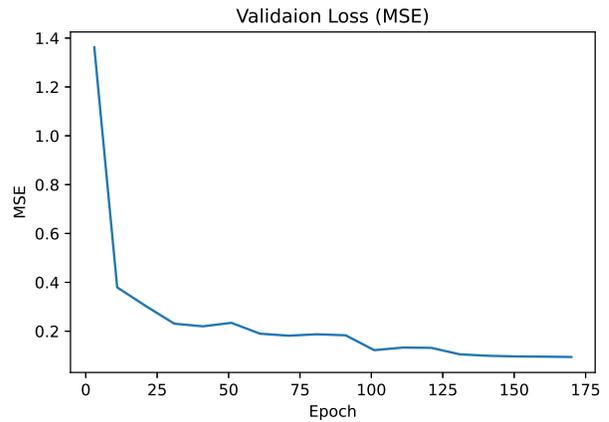}
    \caption{Graph of MSE vs Epochs}
    \label{fig:my_label}
\end{figure}

\begin{figure}[H]
    \centering
    \includegraphics[height=6cm]{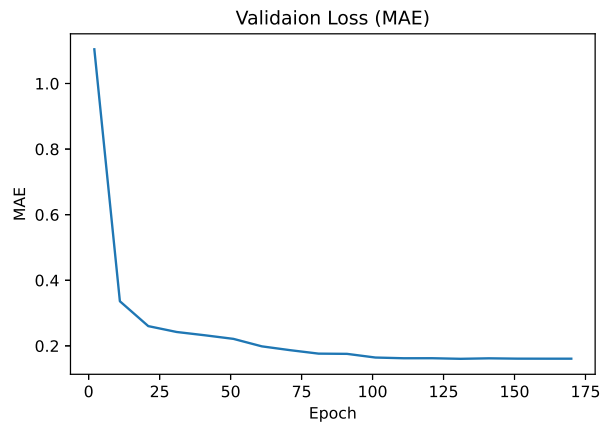}
    \caption{Graph of MAE vs Epochs}
    \label{fig:my_label}
\end{figure}

We compared our result with some contemporary works related to ventilator pressure prediction. Our proposed model performed better than those research's proposed models. The following table shows a comparison of our work with some previous research. The same dataset was used in all of these research works.

\begin{table}[H]
\centering
\caption{Comparing our result with some contemporary works}\label{tab1}
\begin{tabular}{|l|l|l|}
\hline
Authors & Techniques  & MAE \\
\hline
Abdelghani Belgaid\cite{ds} & ResBiLSTM & 0.15\\
\hline
Wadne et al.\cite{vinod} & RNN &  0.3256 (average) \\
\hline
Our research & Bi-LSTM and Bi-GRU  & 0.145\\
\hline
\end{tabular}
\end{table}

We took the average of the MAE of five samples because research by Wadne et al.\cite{vinod} showed results for five samples. 

\section{Conclusion}
A methodology was designed and implemented successfully to simulate a ventilator and predict a ventilator's pressure. The main deep learning networks used in this research were Bi-LSTM and Bi-GRU, which are different kinds of recurrent neural networks. The proposed model was evaluated using MAE and MSE to know the capability of the real-world application of this model. As a result, we found a low error rate in our proposed model. However, this model also has shortcomings, such as not being tested on a real-time mechanical ventilator. Therefore, we plan to decrease the error rate and make an entire system so that it can be used directly in the real-world ventilator of a hospital.

\bibliographystyle{unsrt} 
\bibliography{conference_041818} 

\begin{thebibliography}{10}

\bibitem{me-vent}
Claude Gu{\'e}rin and Patrick L{\'e}vy.
\newblock Easier access to mechanical ventilation worldwide: an urgent need for
  low income countries, especially in face of the growing covid-19 crisis.
\newblock {\em European Respiratory Journal}, 55:6, 2020.

\bibitem{mo}
M.~Yang and J.~Wang.
\newblock Adaptability of financial time series prediction based on bilstm.
\newblock {\em Procedia Computer Science}, 199:18--25, 2022.

\bibitem{rong}
R.~Liang, X.~Chang, P.~Jia, and C.~Xu.
\newblock Mine gas concentration forecasting model based on an optimized bigru
  network.
\newblock {\em ACS omega}, 5(44):28579--28586, 2020.

\bibitem{zafar}
N.~Zafar, I.~U. Haq, J.~U.~R. Chughtai, and O.~Shafiq.
\newblock Applying hybrid lstm-gru model based on heterogeneous data sources
  for traffic speed prediction in urban areas.
\newblock {\em Sensors}, 22(9):3348, 2022.

\bibitem{NP}
Nilesh~P. Sable et~al.
\newblock Pressure prediction system in lung circuit using deep learning.
\newblock In {\em ICT with Intelligent Applications. , Singapore}, pages
  605--615. 2023.

\bibitem{lstm}
S.~Hochreiter and J.~Schmidhuber.
\newblock Long short-term memory.
\newblock {\em Neural computation}, 9(8):1735--1780, 1997.

\bibitem{l-eqn}
Sima Siami-Namini, Neda Tavakoli, and Akbar~Siami Namin.
\newblock The performance of lstm and bilstm in forecasting time series.
\newblock In {\em 2019 IEEE International Conference on Big Data (Big Data)}.
  IEEE, 2019.

\bibitem{b_pic}
Rabah Alzaidy, Cornelia Caragea, and C.~Lee Giles.
\newblock {\em Bi-LSTM-CRF sequence labeling for keyphrase extraction from
  scholarly documents}.
\newblock The world wide web conference, 2019.

\bibitem{cho}
K.~Cho, B.~Van~Merri{\"e}nboer, D.~Bahdanau, and Y.~Bengio.
\newblock On the properties of neural machine translation: Encoder-decoder
  approaches. arxiv.
\newblock preprint, 2014.

\bibitem{gru2}
J.~Chung, C.~Gulcehre, K.~Cho, and Y.~Bengio.
\newblock Empirical evaluation of gated recurrent neural networks on sequence
  modeling. arxiv.
\newblock preprint, 2014.

\bibitem{bgr1}
Qing Zhu et~al.
\newblock A hybrid vmd--bigru model for rubber futures time series forecasting.
\newblock {\em Applied Soft Computing}, 84, 2019.

\bibitem{bgr2}
Rong Liang et~al.
\newblock Mine gas concentration forecasting model based on an optimized bigru
  network.
\newblock {\em ACS omega}, 5(44):28579--28586, 2020.

\bibitem{selu1}
G.~Klambauer, T.~Unterthiner, A.~Mayr, and S.~Hochreiter.
\newblock Self-normalizing neural networks.
\newblock {\em Advances in neural information processing systems}, 30, 2017.

\bibitem{robust}
V.~G. Raju, K.~P. Lakshmi, V.~M. Jain, A.~Kalidindi, and V.~(2020 Padma.
\newblock August).
\newblock In InThird International, editor, {\em Study the influence of
  normalization/transformation process on the accuracy of supervised
  classification}, pages 729--735. Conference on Smart Systems and Inventive
  Technology (ICSSIT) . IEEE, 2020.

\bibitem{ds}
Abdelghani Belgaid.
\newblock {\em Deep Sequence Modeling for Pressure Controlled Mechanical
  Ventilation}.
\newblock medRxiv, 2022.

\bibitem{vinod}
Dr.~Vinod Wadne et~al.
\newblock Pressure prediction system in lung circuit using deep learning and
  machine learning.
\newblock {\em International Research Journal of Engineering and Technology
  (IRJET), 0}, 9, May 2022.

\end{thebibliography}
\end{document}